\documentclass[sigconf]{aamas} 
\usepackage{balance} % for balancing columns on the final page
\usepackage{graphicx}
\usepackage{algorithm}
\usepackage{algorithmic}
\usepackage{tikz}
\usetikzlibrary{positioning, shapes, arrows}

\newcommand {\defn} {\triangleq}

\renewcommand \Pr {\mathop{\mbox{\ensuremath{\mathbb{P}}}}\nolimits}
\newcommand \pol {\pi}

\let\Pr\relax
\newcommand \Pr {\mathbb{P}}

\DeclareMathAlphabet{\mathpzc}{OT1}{pzc}{m}{it}

\if 1

\else

\fi

\newcommand\hecomment[1]{{\color{red}{[HE: \texttt{#1}]}}}

\usepackage{caption,subcaption}
\usepackage{amsmath,array}

\usepackage{mathtools}
\usepackage{bm}
\usepackage{graphics}
\usepackage[mode=buildnew]{standalone}
\usepackage{hyperref}
\setcopyright{ifaamas}
\acmConference[AAMAS '22]{Proc.\@ of the 21st International Conference
on Autonomous Agents and Multiagent Systems (AAMAS 2022)}{May 9--13, 2022}
{Auckland, New Zealand}{P.~Faliszewski, V.~Mascardi, C.~Pelachaud,
M.E.~Taylor (eds.)}
\copyrightyear{2022}
\acmYear{2022}
\acmDOI{}
\acmPrice{}
\acmISBN{}

\acmSubmissionID{???}

\title[Risk-Sensitive Bayesian Games]{Risk-Sensitive Bayesian Games for Multi-Agent Reinforcement Learning under Policy Uncertainty}

\author{Hannes Eriksson}
\affiliation{
  \institution{Zenseact AB}
  \city{Gothenburg}
  \country{Sweden}}
\email{hannes.eriksson@zenseact.com}

\author{Debabrota Basu}
\affiliation{
  \institution{Scool, INRIA Lille- Nord Europe}
  \city{Lille}
  \country{France}}

\author{Mina Alibeigi}
\affiliation{
  \institution{Zenseact AB}
  \city{Gothenburg}
  \country{Sweden}}

\author{Christos Dimitrakakis}
\affiliation{
  \institution{University of Oslo}
  \city{Oslo}
  \country{Norway}}

\begin{abstract}
In stochastic games with incomplete information, the uncertainty is evoked by the lack of knowledge about a player's own and the other players' types, i.e. the utility function and the policy space, and also the inherent stochasticity of different players' interactions. In existing literature, the risk in stochastic games has been studied in terms of the inherent uncertainty evoked by the variability of transitions and actions. In this work, we instead focus on the risk associated with the \textit{uncertainty over types}. We contrast this with the multi-agent reinforcement learning framework where the other agents have fixed stationary policies and investigate risk-sensitiveness due to the uncertainty about the other agents' adaptive policies.
We propose risk-sensitive versions of existing algorithms proposed for risk-neutral stochastic games, such as Iterated Best Response (IBR), Fictitious Play (FP) and a general multi-objective gradient approach using dual ascent (DAPG). Our experimental analysis shows that risk-sensitive DAPG performs better than competing algorithms for both social welfare and general-sum stochastic games.
\end{abstract}


\newcommand{\BibTeX}{\rm B\kern-.05em{\sc i\kern-.025em b}\kern-.08em\TeX}

\begin{document}

%%% The following commands remove the headers in your paper. For final 
%%% papers, these will be inserted during the pagination process.

\pagestyle{fancy}
\fancyhead{}

%%% The next command prints the information defined in the preamble.

\maketitle 

%%%%%%%%%%%%%%%%%%%%%%%%%%%%%%%%%%%%%%%%%%%%%%%%%%%%%%%%%%%%%%%%%%%%%%%%

\section{Introduction}

% https://arxiv.org/pdf/2104.04901.pdf
%https://openreview.net/pdf?id=wfZGut6e09
%%% What is a Bayesian game
A Bayesian game~\citep{harsanyi1967games} is a game of incomplete information in which multiple players or agents act simultaneously and the corresponding uncertainty is modelled using Bayesian posterior updates. 
%%% What is the goal of a Bayesian game
The goal of the game is to identify a set of \emph{strategies} or \emph{policies} that determine how each of the agents will interact with the environment while optimising the corresponding utility functions. We delineate between \emph{pure}, i.e. deterministic, and \emph{mixed}, i.e. stochastic, strategies. Upon fixing the set of strategies that the agents will use, the game can be evaluated and the utility of each of the players can be measured. 
%%% Applications of Bayesian games?
Bayesian game formalism has been used to model threats in the defence industry~\citep{brynielsson2004bayesian}, autonomous driving~\citep{peng2019bayesian}, intrusion detection in wireless networks~\citep{liu2006bayesian}, auctions~\citep{naroditskiy2007using} and more.

Depending on what their individual goals are, the players could act in a cooperative~\citep{oliehoek2010heuristic}, adversarial~\citep{huang2019dynamic} or individual manner.
In the case where all of the players have identical utility functions, we call the game \emph{cooperative}. If instead, the goal of all the agents is to maximise the joint utility, we term the game a \emph{social welfare} game. Further, some agents could attempt to minimise the utility of others, in that case, we have an \emph{adversarial} game. Finally, if there is no restriction on the utility functions,  we have a \emph{general} game. 
%As such, the choice of utility functions is imperative to model the Bayesian game depending on an application.
The flexibility of the framework of the Bayesian game allows for the optimisation of any of these goals by specifying utility functions for each of the agents. 
Thus, the risk-sensitive Bayesian game formulation and the corresponding algorithms that we propose are independent of the choice of the utility function. Due to brevity of space, we perform experimental evaluations only for social welfare and general games.

\iffalse
In this work, we are investigating risk-sensitive approaches to multi-agent reinforcement learning (MARL) where agents interact with each other in the same environment. 
The motivation for this is multifold, by leveraging MARL techniques and explicitly modelling other agents in the environment, we can investigate the risk that arises due to the uncertainty of the other agents' behaviors. Further, many interesting settings involve multiple agents, such as autonomous driving. In autonomous driving, the dynamics of the system are typically known (how the different actuators affect vehicle motion, etc.), whereas the interaction with other vehicles is more unclear. 
The decision-making process of the ego vehicle ultimately depends on how other vehicles will act (and interact) with the ego vehicle.
\fi

\noindent\paragraph{Related Works: Formulations.}
%%% Bayesian games as MARL
A line of research reduces multi-agent reinforcement learning (MARL) to the stochastic game setting where the agents aim to maximise their own utility functions~\citep{chalkiadakis2003coordination, heinrich2016deep, lanctot2017unified}. Under the assumption that all players except one have fixed stationary strategies, the problem is formulated as a partially observable Markov decision process (POMDP) and solved using traditional reinforcement learning (RL) techniques. 

%%% Bayesian games with Risk
\citet{mansour2021bayesian} studied Bayesian games with a focus on incentivising exploration which induces risk-seeking behaviour in order to learn more about the task at hand. \citet{tian2021bounded} investigated the risk inherent to a stochastic game and provide algorithms for risk-averse agents. The risk studied in~\citep{capponi2017risk,tian2021bounded} is more akin to \emph{aleatory}, i.e. inherent, risk encountered in RL previously~\citep{mihatsch2002risk}.

In contrast, our work considers the risk associated the incomplete information about types, i.e. the utility function and policy space. 
%%% RL risk connection with games
The risk in Bayesian games due to type uncertainty is connected to the works in RL considering \emph{epistemic} risk~\citep{depeweg2018decomposition, eriksson2020epistemic} as the Bayesian game can be considered a multi-task RL problem where multiple agents are optimised simultaneously. In the case with only one adaptive agent, this reduces to doing epistemic risk-sensitive RL.

%%% The algorithms
\paragraph{Related Works: Algorithms.} A multitude of approaches are proposed to optimise the behaviour of the agents in Bayesian games. One of the approaches is \emph{Iterated Best Response} (IBR), where the algorithm optimises every player's strategies iteratively by selecting the best response for one agent given all the other agents. \citet{ho1998iterated} used this approach to study problems where multiple developers are to construct resorts near natural attractions, where each developer is acting egotistically. In~\citep{naroditskiy2007using}, the authors apply IBR to identify Bayesian Nash equilibria in auctions. The obtained best-response strategy in this setting is identical among bidders. 
% In cases where objectives among agents are opposed, this approach may lead to cycle of optimal policies $\pol_1^1 \rightarrow \pol_2^1 \rightarrow \pol_1^2 \rightarrow \pol_2^2 \rightarrow \pol_1^1 \rightarrow \hdots$, where $\pol_1^1 \succ \pol_2^2 \succ \pol_1^2 \succ \pol_2^1 \succ \pol_1^1$.

Another approach is \emph{Fictitious Play} (FP)~\citep{brown1951iterative}, where one agent is playing fictitiously against the other. Typically, this is done by letting one agent play against a combination of the other agent's past behaviour. For instance, assuming the other agent's strategy is an average over all of the actions it has taken so far. \citet{shapley1964some} shows that for problems with more players and more strategies per player, as in a Bayesian game, there can exist solutions without convergence. But in cooperative setting, the fictitious play demonstrated convergence guarantees~\citet{monderer1996fictitious}. In this work, we use a variant of FP that is a rolling average over the past weights of the other agent. 
% The addition of this fictitious agent may resolve potential cycling issues as one agent will adapt more smoothly.

Finally, one can simply optimise all agents simultaneously by updating all of their strategies by ascending according to each of their objectives. This approach (for the two-player case) is called \emph{Dual Ascent Policy Gradient} (DAPG) henceforth in this work. Multi-objective policy gradients have been studied before in the field of MARL,~\citet{lu2021decentralized} propose a safe gradient descent ascent approach for multiple decentralised agents. A crucial difference between their and our work is the MARL setting, while our work is in a Bayesian game. Further, the risk studied in their work is the inherent (aleatory) risk. Another recent contribution by~\citet{kyriakis2021pareto} involve multi-objective policy gradients used to identify Pareto optimal solutions with unknown and linear utility functions for RL. The setting is notably different as our work focuses on the Bayesian game setting. Likewise, if the problems is cast as a multi-task setting, something that does not apply to the general Bayesian game setting, then~\citet{sener2018multi} propose an applicable solution to identify Pareto optimal solutions using gradient-based multi-objective learning.

There exist also attempts to unify these frameworks, for instance, ~\citet{lanctot2017unified} generalises previous approaches, such as IBR, FP and more to a general framework. From this, they propose a new algorithm, \emph{Policy-Space Response Oracles}, which constructs a meta-strategy taking all of the players' utilities into account. Another work, by~\citet{heinrich2016deep} develop \emph{Neural Fictitious Self-Play}, an extension of \emph{Fictitious Self-Play}~\citep{heinrich2015fictitious}, where agents play against themselves, fictitiously. These strategies are then compared to classical RL algorithms.

\noindent\textit{Contribution:}
We propose a novel framework for managing risk due to type uncertainty in Bayesian games. To the best of our knowledge, this has not been done before. We relate the risk that arises due to type uncertainty to the risk of not knowing the policies of the other players in MARL. Finally, we demonstrate the superiority of the framework of handling this risk experimentally by introducing three risk-sensitive versions of commonly used algorithms for solving stochastic games and evaluating them on toy problems. These are risk-sensitive iterated best response (RS-IBR), risk-sensitive fictitious play (RS-FP) and risk-sensitive dual ascent policy gradient (RS-DAPG).

\section{Background}

A Bayesian game can be defined in a multitude of ways. In this work, we adopt the framework given by~\citet{harsanyi1967games} where a set of players act in an imperfect information game with a common prior over types. 
\begin{definition}[Bayesian Game]
    A Bayesian game can be formulated as $\mathcal{G} = (N, K, \mathcal{S}, \mathcal{A}^N, \mathcal{R}^{N \times K}, \mathcal{T}, \xi, \gamma)$, where $N$ is the number of players, $K$ the number of possible types, $\mathcal{S}$ the state space of the problem, $\mathcal{A}_i$ the action space of player $i$, $\mathcal{R}_i^j$ the reward function of player $i$ for type $j$, $\mathcal{T} : \mathcal{S} \times \mathcal{A}^N \rightarrow \Delta(\mathcal{S})$ the transition distribution of the problem, $\xi$ a common prior over types and $\gamma$ a common discount factor.
\end{definition}

In particular, a Bayesian game $\mathcal{G}$ is equipped with a set of \emph{types}, $\tau_1, \tau_2$, where each type uniquely specifies the utility function and behaviour of a player. Thusly, part of the uncertainty associated with a Bayesian game comes from each player not knowing which type they will be and what type the other players will be. An agent must then prepare itself for all possible player and type configurations and arrive at strategies that can perform well in each scenario.

To evaluate a game $\mathcal{G}$, it needs to commit to a set of policies $\Pi^{N \times K}$, one policy per player and type combination. We denote the policies by $\{\pol_1^1, \hdots, \pol_1^K, \pol_2^1, \hdots, \pol_2^K\} \in \Pi$ for a two-player game with $K$ types. The expected utility $U_1^{j,k}$ associated with player $1$, when player $1$ has type $j$ and player $2$ has type $k$, can be defined in the following way,

\begin{equation}\label{eq:game_utility}
    U_1^{j, k}(\mathcal{G}) \defn \mathbb{E}_{\mathcal{G}}^{\pol_1^j, \pol_2^k}\Big[\sum_{t=0}^\infty \gamma^t r_{1,t}^j \, | \, s_0=s\Big].
\end{equation}

Note that the utility associated with type $j$ for player $1$ depends not only on the policy of player $2$ for type $j$ but for all combinations of types. Combining all the utilities for a particular player by marginalising over the prior $\xi$ can be done in the following way,

\begin{definition}[Expected Utility of a game $\mathcal{G}$]
    The expected utility marginalised over the prior $\xi$ for a given game $\mathcal{G}$ is 
    \begin{align}
        U_1, U_2 &\defn \mathbb{E}_\xi\Big[U_1^{j, k}(\mathcal{G}), U_2^{j,k}(\mathcal{G})\Big]\\
        &= \sum_{j=1}^K \sum_{k=1}^K\xi(\tau_1 = j, \tau_2 = k) (U_1^{j, k}(\mathcal{G}), U_2^{j, k}(\mathcal{G})).
    \end{align}
\end{definition}
\section{Risk in Bayesian games}

In this work, the risk associated with the uncertainty about types is studied. That is, $\xi$, combined with the utility of each of the individual players for that particular type configuration. $U_1^{., .}$ and $U_2^{., .}$ are thus discrete probability distributions with mass equal to $\xi(\tau_1 = j, \tau_2 = k)$ and values $U_1^{j, k}$ and $U_2^{j, k}$. By considering all possible combinations of types we get the following probability mass function,

\begin{equation}
    p_{U_1}(U) = \begin{cases}
        U_1^{1, 1}, &\xi(\tau_1 = 1, \tau_2 = 1)\\
        U_1^{1, 2}, &\xi(\tau_1 = 1, \tau_2 = 2)\\
        &\vdots\\
        U_1^{K, K}, &\xi(\tau_1 = K, \tau_2 = K),
    \end{cases}
\end{equation}  
and similarly for the other agent, $p_{U_2}(U)$.  In this work, we consider conditional value-at-risk (CVaR), studied for instance by~\citet{artzner1999coherent}, which is a risk measure that focuses on the $\alpha\%$ worst-case outcomes of the distribution. CVaR can be defined as follows, $CVaR_\alpha(U) \defn \mathbb{E}[U \, | \, U \leq \nu_\alpha \land \Pr(U \geq \nu_a) = 1-\alpha]$.

Earlier works~\citep{eriksson2021sentinel, rigter2021risk} have shown that coherent risk measures, like CVaR, can be written as perturbed distribution functions, and expectation is taken w.r.t the perturbed distribution yields risk-adjusted utilities.

\iffalse
All coherent risk measures are by definition [by the axioms] Choquet integrals and can be written as \emph{distortion functions}, see~\citep{eriksson2021sentinel, rigter2021risk}. 

\begin{definition}[Distortion function]
    A concave function $U : [0, 1] \rightarrow [0, 1]$ where $U(0) = 0$ and $U(1) = 1$ is called a \emph{distortion function}.
\end{definition}

Because of the subadditivity of $U$, the total risk can thus be upper bounded by the sum of the individual risks, and optimisation for a risk-sensitive game can be taken as a new weighted sum of individual risks.
\fi
\section{Algorithms}

In this section, we describe the basis of the risk-sensitive algorithms proposed in this work. They are, IBR, FP and DAPG with respective risk-sensitive versions RS-IBR, RS-FP and RS-DAPG. All algorithms use softmax policies 
\begin{equation}
    \pi_1^j(a \, | \, s) = \frac{e^{\theta_{1, s}^{j, a}}}{\sum_{a' \in \mathcal{A}_1} e^{\theta_{1, s}^{j, a'}}},
\end{equation}

for type $\tau_1 = j$ and player $1$. $\theta \in \mathbb{R}^{N \times K \times |\mathcal{S}| \times |\mathcal{A}|}$ is thus the full parameter set.

\subsection{Iterated Best Response}

IBR~\citep{ho1998iterated, naroditskiy2007using,bopardikar2017convergence} is an algorithm that iteratively identifies the optimal policy for a given player, keeping the policies of all other players fixed. Since IBR also only updates a subset of the available parameters at each iteration, it might have convergence problems for the general setting. The IBR procedure can be seen in Algorithm~\ref{alg:ibr_fp}.

\subsection{Fictitious Play}

One application of the FP~\citep{brown1951iterative, shapley1964some, monderer1996fictitious} framework can be seen in Algorithm~\ref{alg:ibr_fp}. The setup is similar to that of IBR but with the important distinction that the objective is taken w.r.t a rolling average of the previous parameters of agent one for agent two, where $\bar{\theta}_{1,i+1} \defn \frac{1}{i+1}\sum_{j=0}^{i+1} \theta_{{1, j}}$.

\subsection{Dual Ascent Policy Gradient}

The final algorithm uses the utilities from multiple objectives to compute a gradient step over the whole parameter set. As there are only two agents, with one objective respectively, the algorithm is termed dual ascent and is ascending simultaneously using both objectives. Note that the gradient $\nabla_{\theta_{1}^j} [U_1 \, | \, \mathcal{G}, \theta]$ is non-zero for the type selections $\{\tau_1 = j, \tau_2 = 1 \hdots K\}$ for both objective $U_1$ and $U_2$. The algorithm is displayed in Algorithm~\ref{alg:mopg}.

\begin{algorithm}[t!]
	\caption{Risk-Sensitive Iterated Best Response/Fictitious Play (RS-IBR/FP)}
	\label{alg:ibr_fp}
	\begin{algorithmic}[1]
	    \STATE \textbf{input }: Game $\mathcal{G}$, learning rates $\eta_1, \eta_2$, risk measure $\rho_\alpha$
	    \FOR{$i=0 \hdots \, $ convergence}
	        \FOR{$t=0 \hdots \, $ convergence}
	            \STATE $\theta_{1, i}^{t+1} = \theta_{1, i}^{t} + \eta_1\nabla_{\theta_{1}}\big[\rho_\alpha(U_1) \, | \, \mathcal{G}, \theta_{1, i}^t, \theta_{2, i}\big]$
                %\STATE $\theta_{\Pi_{1, i+1}^{t+1}} = \theta_{\Pi_{1, i}^t} + \eta_1\nabla_{\theta_{\Pi_{1}}}\big[\rho_\alpha(U_1) \, | \, \mathcal{G}, \Pi_{1,i}^t, \Pi_{2, i}^\infty\big]$

	        \ENDFOR
	        \STATE $\theta_{1, i+1} = \theta_{1, i}^t$
	        \FOR{$t=0 \hdots \, $ convergence}
	            \IF{IBR}
	            \STATE $\theta_{2, i}^{t+1} = \theta_{2, i}^{t} + \eta_2\nabla_{\theta_{2}}\big[\rho_\alpha(U_2) \, | \, \mathcal{G}, \theta_{1, i+1}, \theta_{2, i}^t\big]$
                %\STATE $\theta_{\Pi_{2, i+1}^{t+1}} = \theta_{\Pi_{2, i}^{t}} + \eta_2\nabla_{\theta_{\Pi_{2}}}\big[\rho_\alpha(U_2) \, | \, \mathcal{G}, \Pi_{1, i+1}^\infty, \Pi_{2, i}^t\big]$
                \ENDIF
                \IF{FP}
	            \STATE $\theta_{2, i}^{t+1} = \theta_{2, i}^{t} + \eta_2\nabla_{\theta_{2}}\big[\rho_\alpha(U_2) \, | \, \mathcal{G}, \bar{\theta}_{1, i+1}, \theta_{2, i}^t\big]$
                \ENDIF
            \ENDFOR
            \STATE $\theta_{2, i+1} = \theta_{2, i}^t$
	    \ENDFOR
	\end{algorithmic}
\end{algorithm}

\begin{algorithm}[t!]
	\caption{Risk-Sensitive Dual Ascent Policy Gradient (RS-DAPG)}
	\label{alg:mopg}
	\begin{algorithmic}[1]
	    \STATE \textbf{input }: Game $\mathcal{G}$, learning rates $\eta_1, \eta_2$, risk measure $\rho_\alpha$
	    \FOR{$i=0 \hdots \, $ convergence}
	        \FOR{$j=1 \hdots K$}
    	        \STATE $\theta_{1, i+1}^j = \theta_{1, i}^j + \eta_1\nabla_{\theta_{1}^j}\Big( \big[\rho_\alpha(U_1) \, | \, \mathcal{G}, \theta_i\big]+\big[\rho_\alpha(U_2) \, | \, \mathcal{G}, \theta_i\big]\Big)$
    	        %\STATE $\theta_{\pol_1^j} = \theta_{\pol_1^j} + \eta_1\nabla_{\theta_{\pol_1^j}}\Big( \big[\rho_\alpha(U_1) \, | \, \mathcal{G}, \Pi_i\big]+\big[\rho_\alpha(U_2) \, | \, \mathcal{G}, \Pi_i\big]\Big)$
    	        \STATE $\theta_{2, i+1}^j = \theta_{2, i}^j + \eta_2\nabla_{\theta_{2}^j}\Big( \big[\rho_\alpha(U_1) \, | \, \mathcal{G}, \theta_i\big]+\big[\rho_\alpha(U_2) \, | \, \mathcal{G}, \theta_i\big]\Big)$
	        \ENDFOR
	    \ENDFOR
	\end{algorithmic}
\end{algorithm}
\section{Experiments}

\begin{table}[ht]
    \centering
    \caption{Social welfare setting, evaluated over $100$ toy experiments. The first column indicates the algorithm evaluated. If the algorithm is prefixed by \texttt{RS-}, then it uses a risk-sensitive objective. The second column is the social welfare objective $(U_1+U_2)/2$. An algorithm with higher social welfare performed better. The third column displays the risk-adjusted social welfare objective, using CVaR with threshold parameter $\alpha=0.25$. All results are averaged over $100$ different games and the mean and standard deviation is reported.}
    \begin{tabular}{lcc}\toprule
         \textit{Agent} & \textit{Social Welfare} $\pm\sigma$ & \textit{CVaR}$_{\alpha=0.25}$ \textit{Social Welfare} $\pm\sigma$\\ \midrule
         MMBI~\citep{dimitrakakis:mmbi:ewrl:2011}& $5.69\pm3.10$& $-5.54\pm10.56$ \\
         IBR~\citep{ho1998iterated, naroditskiy2007using,bopardikar2017convergence}& $2.57\pm2.65$& $-1.10\pm7.30$ \\
         FP~\citep{brown1951iterative, shapley1964some,monderer1996fictitious}& $2.60\pm2.64$ & $-1.11\pm7.35$\\
         DAPG & $\mathbf{6.01\pm3.04}$ & $-9.40\pm9.58$\\
         RS-MMBI~\citep{eriksson2020epistemic} & $-4.41\pm3.97$ & $22.84\pm14.02$\\
         RS-IBR & $-1.79\pm3.10$ & $12.67\pm10.13$\\
         RS-FP & $-1.77\pm3.11$ & $12.6\pm10.17$\\
         RS-DAPG & $-4.21\pm3.13$ & $\mathbf{23.73\pm13.18}$ \\\bottomrule
    \end{tabular}
    \label{tab:social}
\end{table}

\begin{table*}[ht]
    \centering
    \caption{General setting, evaluated over $100$ toy experiments. The first column indicates the algorithm evaluated. If the algorithm is prefixed by \texttt{RS-}, then it uses a risk-sensitive objective. The second and third columns are the individual utilities of the agents, $U_1, U_2$. The fourth and fifth columns display the risk-adjusted individual utility for the respective player, using CVaR with threshold parameter $\alpha=0.25$. All results are averaged over $100$ different games and the mean and standard deviation is reported.}
    \begin{tabular}{lcccc}\toprule
         \textit{Agent} & \textit{Utility}$_1 \pm\sigma$ & \textit{Utility}$_2 \pm\sigma$ & \textit{CVaR}$_{\alpha=0.25}$ \textit{Utility}$_1 \pm\sigma$ & \textit{CVaR}$_{\alpha=0.25}$ \textit{Utility}$_2 \pm\sigma$\\ \midrule
         MMBI~\citep{dimitrakakis:mmbi:ewrl:2011}& $5.50\pm5.03$& $5.88\pm5.31$& $-3.73\pm14.93$& $-2.64\pm17.16$\\
         IBR~\citep{ho1998iterated, naroditskiy2007using,bopardikar2017convergence}& $3.98\pm3.92$& $0.13\pm3.81$& $-3.50\pm11.25$& $8.67\pm12.93$ \\
         FP~\citep{brown1951iterative, shapley1964some,monderer1996fictitious}& $3.86\pm3.78$ & $0.21\pm3.74$& $-3.16\pm9.89$& $8.72\pm12.84$\\
         DAPG & $\mathbf{5.93}\pm\mathbf{4.74}$ & $\mathbf{6.09}\pm\mathbf{4.71}$& $-6.19\pm14.13$& $-5.81\pm14.04$\\
         RS-MMBI~\citep{eriksson2020epistemic}& $-4.45\pm6.11$& $-4.37\pm5.66$& $22.48\pm18.15$& $24.63\pm22.04$\\
         RS-IBR & $-2.74\pm4.90$ & $0.68\pm4.24$& $18.04\pm14.88$& $6.19\pm12.14$\\
         RS-FP & $-1.98\pm4.87$ & $2.43\pm3.09$& $18.35\pm9.99$& $1.30\pm8.70$\\
         RS-DAPG & $-4.76\pm5.80$ & $-4.50\pm5.37$& $\mathbf{25.60}\pm\mathbf{18.47}$& $\mathbf{26.53}\pm\mathbf{22.03}$ \\ \bottomrule
    \end{tabular}
    \label{tab:general}
\end{table*}

We investigate randomly generated toy problems, where $\mathcal{T}$ is generated from a product of Dirichlet distributions, all with parameter $\alpha = 1$. The individual reward functions for the players and types are also randomly generated, where $\mathcal{R}$ is generated from a product of normal distributions with parameters $\mu = 0, \sigma = 1$. Performance of the algorithms is then evaluated in two distinct settings, one where the social welfare is maximised and the other in a general multi-objective setting.

The metrics of interest in the social welfare setting is the average utility $(U_1 + U_2)/2$ among the players and a risk-adjusted version of it, $CVaR_\alpha[(U_1 + U_2)/2]$, where $\alpha \in (0, 1]$ is a parameter that controls for the sensitivity to risk. Larger values of utility are better and likewise larger risk-adjusted utilities are preferred. 

In the general setting with multiple objectives, we study the individual utilities among the agents and the algorithms with the best performance is the set of non-dominated algorithms in a Pareto optimality sense.

\subsection{Social Welfare}

\balance 
% for AAMAS only, should be on the final page, on left column

A stochastic game consisting of multiple players with a common goal to maximise the social welfare can be re-instantiated as a Bayesian MDP, which can be solved approximately using backward induction, see \textsc{MMBI} in~\citet{dimitrakakis:mmbi:ewrl:2011} for the risk-neutral case and \textsc{RS-MMBI} in~\citet{eriksson2020epistemic} for the risk-sensitive case. The results acquired from evaluating the social welfare setting is summarised in Table~\ref{tab:social}.

\subsection{General Case}

\begin{figure}
    \centering
    \includegraphics[width=0.23\textwidth]{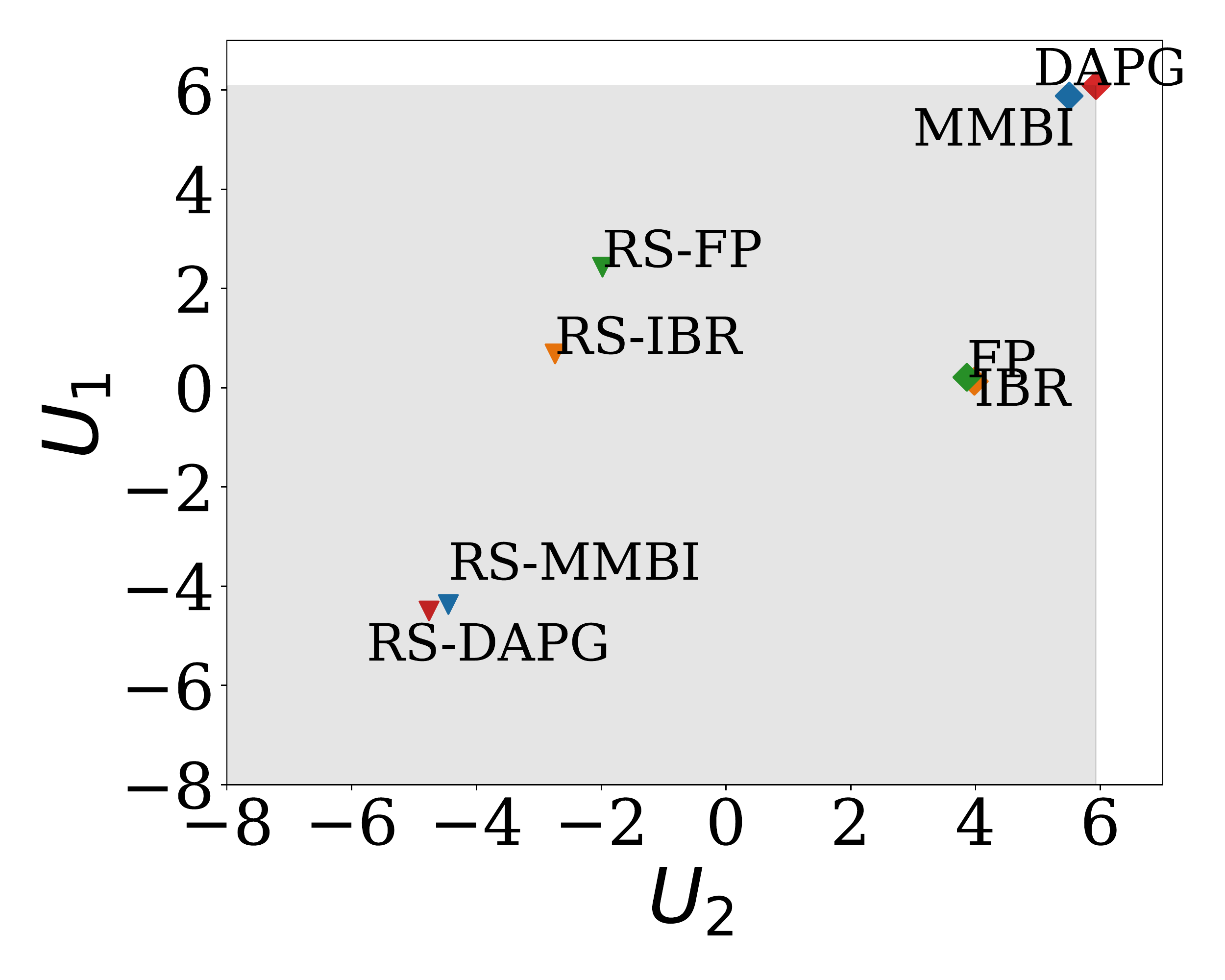}
    \includegraphics[width=0.23\textwidth]{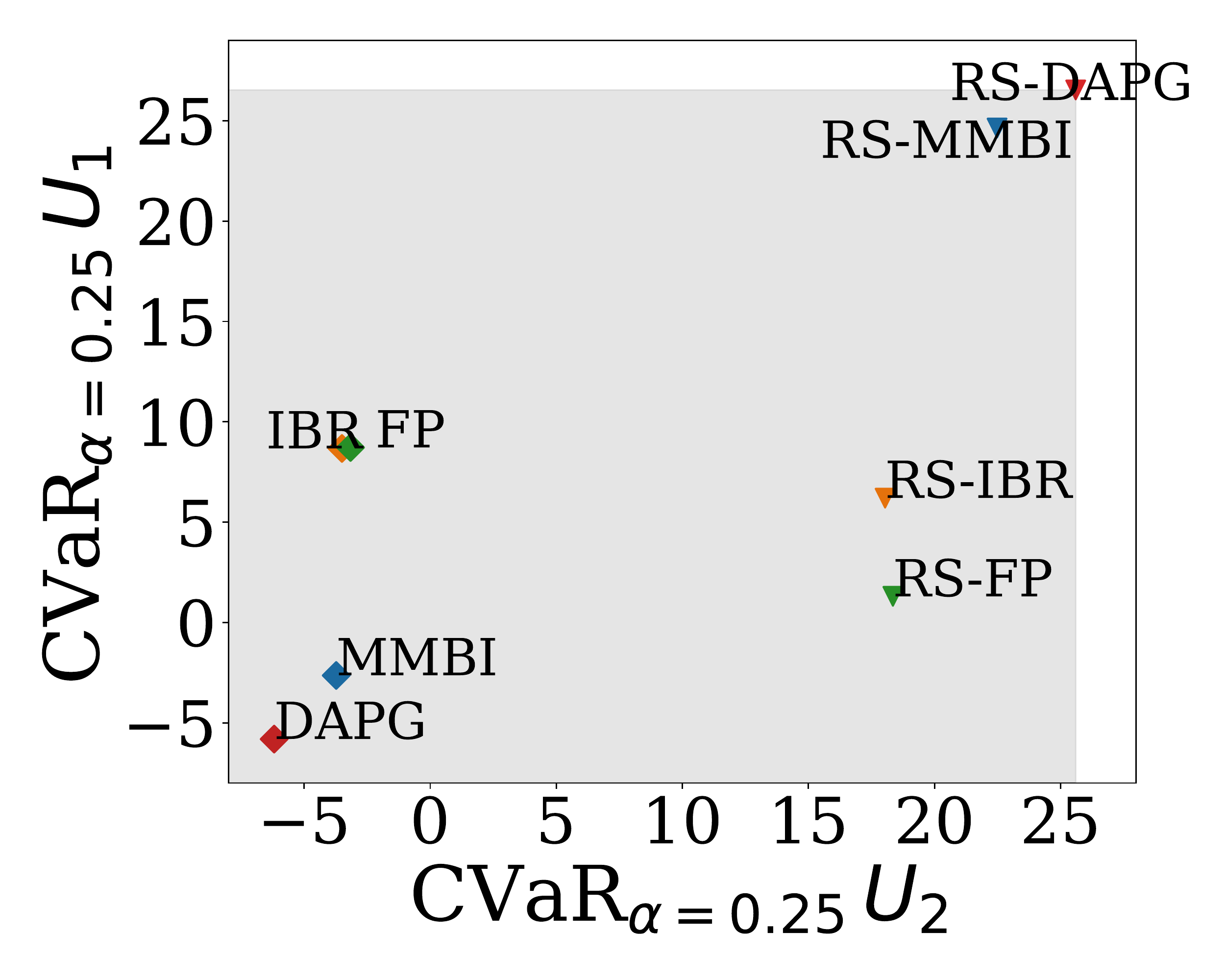}
    \caption{Pareto front the utilities in the toy experiment averaged over $100$ experiments. The shaded region indicates non-domination and the points in the non-shaded region are admissible solutions that are considered equivalent.}
    \label{fig:pareto_front}
\end{figure}

In the general case, there are multiple objectives, one for each agent. We, therefore, study the concept of Pareto optimality~\citep{banerjee2007reaching} to determine which algorithms produce the best results. Figure~\ref{fig:pareto_front} displays the Pareto front by the non-shaded region. The points in the shaded region are all dominated by points in the non-shaded region. The Pareto front of dominating algorithms, $P(\mathbf{A})$, can be defined as,

\begin{equation}
    P(\mathbf{A}) \defn \{A_1 \in \mathbf{A} : \{A_2 \in \mathbf{A} : A_2 \succ A_1, A_1 \neq A_2\} = \emptyset\},
\end{equation}

that is, the set of non-dominated algorithms in $\mathbf{A}$. The Pareto fronts of the risk-neutral and risk-sensitive objectives as seen in Figure~\ref{fig:pareto_front} seem to indicate the general gradient approach in Algorithm~\ref{alg:mopg} is superior to the IBR and FP approaches. In Table~\ref{tab:general} we can see that the risk-sensitive algorithms outperform the risk-neutral algorithms when optimising for the risk-sensitive objective as one would expect. Likewise, for the risk-neutral objective, the risk-neutral algorithms outperform. We can see that the \textit{pure} strategies generated by MMBI and RS-MMBI perform well, although they are bested by the dual ascent approach in both settings. 

%\hecomment{The space of policies is Markov policies, if we had adaptive policies we would probably see a larger difference between MMBI and DAPG.}

\section{Discussion and Future Work}

Extensions of this work would include evaluation of the algorithms in the cooperative settings described in~\citet{dimitrakakis2017multi}, they are, \textbf{Multilane Highway} and \textbf{Food and Shelter} which would be interesting testbeds for the algorithms. Another setting to consider is given by~\citet{tian2021bounded} which is a navigation task evaluated for risk-sensitive games. In their work, they also describe another interesting risk measure.

Proper convergence analysis of the dual ascent policy gradient algorithm for Bayesian games is also something worth looking into as well as the identification of the Pareto optimal set of solutions in order to study Bayesian Nash equilibria in this setting.

%%%%%%%%%%%%%%%%%%%%%%%%%%%%%%%%%%%%%%%%%%%%%%%%%%%%%%%%%%%%%%%%%%%%%%%%

%%% The acknowledgments section is defined using the "acks" environment
%%% (rather than an unnumbered section). The use of this environment 
%%% ensures the proper identification of the section in the article 
%%% metadata as well as the consistent spelling of the heading.

\begin{acks}
This work was partially supported by the Wallenberg AI, Autonomous Systems and Software Program (WASP) funded by the Knut and Alice Wallenberg Foundation and the computations were performed on resources at Chalmers Centre for Computational Science and Engineering (C3SE) provided by the Swedish National Infrastructure for Computing (SNIC).
\end{acks}
%\acks{This work was partially supported by the Wallenberg AI, Autonomous Systems and Software Program (WASP) funded by the Knut and Alice Wallenberg Foundation and the computations were performed on resources at Chalmers Centre for Computational Science and Engineering (C3SE) provided by the Swedish National Infrastructure for Computing (SNIC).}

%%%%%%%%%%%%%%%%%%%%%%%%%%%%%%%%%%%%%%%%%%%%%%%%%%%%%%%%%%%%%%%%%%%%%%%%

%%% The next two lines define, first, the bibliography style to be 
%%% applied, and, second, the bibliography file to be used.

\bibliographystyle{ACM-Reference-Format} 
\bibliography{references}
%\newpage
%\appendix
%\onecolumn
%\input{appendix}

%%%%%%%%%%%%%%%%%%%%%%%%%%%%%%%%%%%%%%%%%%%%%%%%%%%%%%%%%%%%%%%%%%%%%%%%

\end{document}